\documentclass[11pt]{article}
\usepackage[]{acl}

\usepackage{times}
\usepackage{latexsym}
\usepackage[T1]{fontenc}
\usepackage[utf8]{inputenc}
\usepackage{microtype}
\usepackage{booktabs}
\usepackage{amsmath}
\usepackage{graphicx}
\usepackage{tikz}
\usetikzlibrary{positioning, arrows.meta, fit, calc, shapes.geometric,
                shadows.blur, backgrounds}

\usepackage{xcolor}
\usepackage{xurl}

\title{From a Word-Level Dictionary to Sentence-Level Semantics: Multilingual Grievance Labelling with Contextual Models}

\author{Lin Tian \\
  Behavioral Data Science \\
  University of Technology Sydney \\
  Sydney, Australia \\
  \texttt{Lin.Tian-3@uts.edu.au} \\\And
  Marian-Andrei Rizoiu \\
  Behavioral Data Science \\
  University of Technology Sydney \\
  Sydney, Australia \\
  \texttt{Marian-Andrei.Rizoiu@uts.edu.au} \\}

\begin{document}
\maketitle

\begin{abstract}
Grievance is one of the warning signs
analysts look for when assessing threats of violence. It is increasingly measured
at scale from online text, most often with word-level lexicons like the Grievance
Dictionary that score by matching weighted terms. Such matching is a fast and transparent proxy, but it cannot resolve
whether a term is asserted, quoted, negated, or condemned. These lexicons are
also often evaluated on pools enriched with the very examples they retrieve, so a
high score partly reflects agreement with the lexicon's own selection rule. Examining a
five-language, 2{,}000-item evaluation pool, we find its halves separated almost
perfectly by the lexicon itself: every item labeled ``random'' is in fact
lexicon-negative, so the lexicon's apparent macro-AUROC of 0.686 collapses to a
0.500 floor fixed by construction. We keep
the dictionary's 22-construct ontology but replace term matching with
context-reading models, evaluated on a non-circular benchmark that separates
unconditional-random, lexicon-positive, and lexicon-negative strata across five
languages. Reading the full post rather than the target sentence alone helps most
where the lexicon is silent, raising average precision on lexicon-negative text
from 0.14 to 0.20, with the largest gains on quoted, implicit, and cross-sentence
grievance. Together, these results show that grievance is measured more faithfully
by reading the surrounding context, and more honestly when tested on text the
lexicon did not select.
We release our code and benchmark at \url{https://github.com/behavioral-ds/multilingual_grievance}.
\end{abstract}

\section{Introduction}

On 19 February 2020, before killing nine people in Hanau, Germany, the attacker
posted a German-language manifesto and videos on his personal website setting out
the grievances he believed justified the act.\footnote{\url{https://www.france24.com/en/20200219-shooting-leaves-several-people-dead-in-western-german-city-of-hanau}}
This kind of leakage is the rule rather than the exception: most lone attackers
commit their grievances to text or recordings before acting, typically in the
oblique register of a warning behavior rather than an explicit threat
\citep{meloy2012warning,gill2014bombing}; the Christchurch attacker, for one,
aired his intentions on anonymous forums for months beforehand
\citep{wilson2024christchurchposts,bogle2026christchurchcomments}. Because
analysts cannot read every post
at scale \citep{brynielsson2013harvesting,cohen2014detecting}, automated
grievance measurement has become an attractive first filter. Yet the language
that matters is context-laden and often adversarial. The Christchurch attacker's
manifesto \citep{royalcommission2020planning} was deliberately layered with irony,
quotation, and internet in-jokes engineered to be taken literally and misread by
anyone---human or machine---scanning it for meaning.\footnote{\url{https://www.bellingcat.com/news/2019/03/15/shitposting-inspirational-terrorism-and-the-christchurch-mosque-massacre/}}

The core difficulty is that the same words carry different force in different
frames. The word \emph{threat} can express a warning (``They threatened us''),
report someone else's speech (``The article quoted the threat''), or reject it
(``We condemn these threats''). A term-matching instrument sees closely related
evidence in all three cases; a reader uses syntax, stance, discourse, and speaker
attribution to distinguish them. The reverse problem is equally important:
``They took everything from me. Tomorrow, they will answer for it'' can express
grievance and menace without using an obvious dictionary term. Should such an
instrument count the terms a text contains, or judge what its author actually
expressed?

These are not edge cases. Grievance, a perceived wrong relevant to professional
threat assessment, is a latent, context-dependent construct, not a bag of words
\citep{meloy2012warning,gill2014bombing}. The Grievance Dictionary
\citep{vandervegt2021grievance} offers an interpretable starting point: a
weighted lexicon of 22 constructs, later translated and psychometrically
evaluated in Dutch, German, and Italian \citep{vandervegt2025translations}. Its
inventory fixes \emph{what} to measure; weighted term occurrence is only one way
to decide \emph{whether the author expressed it}.


Examining an existing five-language evaluation pool, we find its two halves
separated perfectly by the lexical anchor (every ``lexicon-positive'' item
scores at least 0.30, every ``random'' item below 0.05): the second half is a
\emph{lexicon-negative challenge set}, not a random sample, yet three-annotator
consensus still marks constructs in 228 of its items. Being target-only, the
pool also cannot test whether context changes judgments. This raises three
questions: how much does lexicon-conditioned sampling distort what a lexical
score appears to measure (\textbf{RQ1}); how prevalent are grievance constructs
in unconditionally sampled text, and how well are they measured when the
lexicon does not choose the test set (\textbf{RQ2}); and does reading the full
post change judgments and improve models, and where (\textbf{RQ3})?

To answer these questions, we contribute: (i) a \textbf{selection analysis}
quantifying the circularity and the consensus-marked evidence hidden below the
anchor threshold (RQ1); (ii) a \textbf{measurement framework} that keeps the
dictionary's 22-construct ontology, demotes lexical scores to one fallible
instrument, and types its failures by stance, attribution, discourse, and
multilingual realization; (iii) a \textbf{non-circular five-language benchmark}
with an unconditional-random stratum, giving the first base-rate estimate for
these constructs ($12.9\%$), plus a counterbalanced target-only/full-context
subset (RQ2); and (iv) \textbf{matched-pair evidence that context works}: a
full-post encoder beats its target-only twin on every stratum, most ($39\%$
relative average precision) where the lexicon is silent, with gains
concentrated on quoted, implicit, and cross-sentence grievance (RQ3).

\begin{figure*}[t]
\centering
\includegraphics[width=\linewidth]{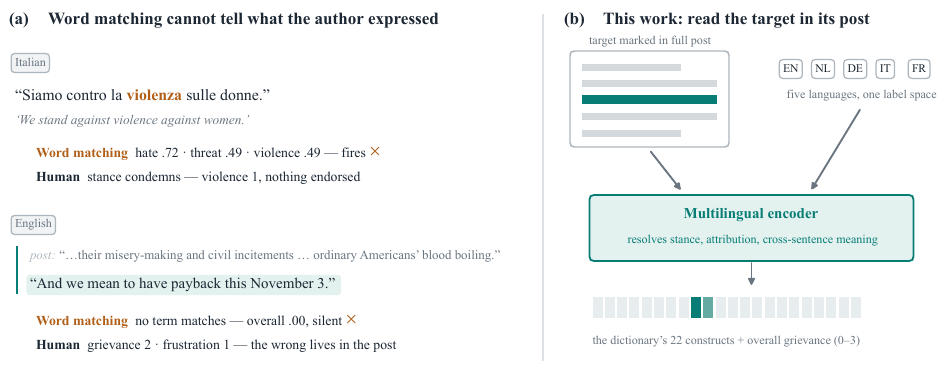}
\caption{\textbf{Measuring grievance by reading context rather than matching
terms.} \textbf{(a)} Two real targets from the five-language pool. An Italian
sentence condemning violence still activates six lexical constructs (three
shown), while an English sentence with anchor score 0 carries a
three-annotator grievance rating of 2 whose perceived wrong is supplied only
by the source post. \textbf{(b)} We keep the Grievance Dictionary's
22-construct ontology but score each construct by reading the marked target
inside its full post, with one multilingual encoder and one shared label space
across English, Dutch, German, Italian, and French. Targets are shortened with
user metadata omitted; their target-only consensus labels are not full-context
adjudications.}
\label{fig:overview}
\end{figure*}

\section{Related Work}

\paragraph{Grievance as lexical measurement.}
Threat-assessment research identifies grievance and related warning behaviors as
potentially relevant signals, while emphasizing that no isolated linguistic
marker proves intent \citep{meloy2012warning,gill2014bombing}. Computational
work has consequently explored linguistic markers of radicalization and threat
\citep{brynielsson2013harvesting,cohen2014detecting}. The Grievance Dictionary
\citep{vandervegt2021grievance} organizes weighted terms into 22 constructs in
the tradition of psychologically grounded resources such as LIWC
\citep{tausczik2010psychological,pennebaker2015development}. Later translations
to Dutch, German, and Italian show the value of multilingual extension but also
uneven reliability, especially for some Italian categories
\citep{vandervegt2025translations}. We retain this construct inventory while
treating its term lists as one fallible measurement instrument rather than
ground truth.

\paragraph{Context and functional evaluation.}
Lexicons are transparent, inexpensive, and reproducible, but bag-of-terms
scores do not model compositional meaning or discourse stance. Related work on
abuse and hate speech demonstrates failures under negation, quotation,
counter-speech, reclaimed language, and speaker attribution
\citep{rottger2021hatecheck,kennedy2020contextualizing}. Functional tests are
particularly useful because aggregate in-domain scores can conceal such
systematic failures. Our focus differs in two ways: grievance spans a broader
construct inventory than hate or offense, and our unit is a target sentence
interpreted inside a marked source post. We further treat the method used to
\emph{sample} the evaluation set as part of the validity argument, since a
lexicon-conditioned pool preferentially exposes the cases the lexicon already
retrieves \citep{hovy2021five}.

\paragraph{Construct validity and subjective annotation.}
Separating a construct from its operationalization follows classical construct
validity: scores are meaningful only relative to the inference supported by the
instrument \citep{cronbach1955construct,messick1995validity}. Sentence-level
grievance is subjective, so replicated annotation and an explicit treatment of
disagreement are necessary \citep{artstein2008inter,krippendorff2004content,
plank2022problem,uma2021learning}. The existing target-only pool stores only the
final consensus of three annotators per item, not their independent ratings, so
its disagreement statistics cannot be reconstructed; our contextual benchmark retains both individual and adjudicated labels by language and
condition.

\paragraph{Multilingual contextual models.}
Multilingual encoders such as mBERT, XLM-R, and mDeBERTa support shared
representations across languages
\citep{devlin2019bert,conneau2020unsupervised,he2023debertav3} and are benchmarked
for cross-lingual generalisation \citep{hu2020xtreme}. Multilingual
abuse and offensive-language benchmarks demonstrate both the value and the
difficulty of cross-lingual evaluation
\citep{basile2019semeval,zampieri2019predicting,mandl2019overview,
pamungkas2019cross}. Our target is not generic toxicity: the 22 grievance
constructs include grievance, fixation, desperation, planning, threat, and
weaponry. A shared contextual encoder is therefore evaluated per language and
per construct, with pooled scores treated as secondary.

\paragraph{LLM supervision and distillation.}
LLMs can provide scalable provisional annotations
\citep{gilardi2023chatgpt,ding2023gptannotation}, and their outputs can supervise
smaller models \citep{hinton2015distilling,sanh2019distilbert,wang2021want,
hsieh2023distilling,li2025learningless}, though such pseudo-labels are noisy and
demand careful selection \citep{song2022noisy}. Sample-wise weighting in knowledge
distillation is itself established \citep{lu2021rwkd}; we evaluate
uncertainty-weighted variants as a secondary training condition
(Appendix~\ref{app:legacy}). In this work, distillation is an implementation
route to a deployable multilingual contextual baseline; human validity and
non-circular evaluation remain the central contributions.

\section{Methodology}
\label{sec:method}

\subsection{From term matching to construct expression}
\label{sec:task}
The Grievance Dictionary bundles two separable objects: a construct inventory
$\mathcal{C}=\{c_1,\ldots,c_{22}\}$ and a lexical rule that scores each construct
from weighted term matches. We retain the inventory and replace the rule. A post
$p=(x_1,\ldots,x_{J_p})$ in language $\ell$ contains a target sentence $x_j$; for
each construct $c$ a system predicts
\[
 y_{jc}=f_c(x_j\mid p,\ell)\in[0,1],
\]
the degree to which the post's author expresses $c$ through $x_j$, read in the
context of the full marked post. The prediction unit is always the target
sentence, so every output stays comparable to the dictionary's; the post enters
only as evidence for resolving stance, attribution, and cross-sentence meaning.
An overall grievance judgment is predicted on the same $0$--$3$ scale by a
separate head, never as a twenty-third construct.

Two properties fix the task. It preserves the ontology: each output maps
one-to-one onto a dictionary construct. It changes the measurement: a listed term
is neither necessary nor sufficient for a positive label, which demotes the
lexical rule from the definition of $y_{jc}$ to one fallible estimator of it.

\subsection{Transferring the 22 constructs across languages}
\label{sec:transfer}
The object that must survive translation is the ontology, not any word list. Each
construct $c\in\mathcal{C}$ is fixed by a single language-independent definition;
the English source definitions are rendered into Dutch, German, Italian, and
French by native speakers with back-translation checks, so an annotator or model
in any language scores against the same construct rather than a locally
re-derived one. The 22 labels therefore form one shared output space: construct
$c$ occupies the same index for every language $\ell$, and no language
introduces, splits, or merges a construct.

Transfer enters each instrument at a different level. The lexical anchor
transfers at the \emph{term} level: its weighted lists are the validated
translations of the Dutch, German, and Italian dictionaries
\citep{vandervegt2025translations} plus an unvalidated French seed, matched in a
language-specific stem space (\S\ref{sec:systems}). The contextual encoders
transfer at the \emph{representation} level: a single multilingual encoder
carries one $22$-dimensional prediction head across all languages, so a label
observed in one language updates the shared construct space rather than a
per-language head. The LLM teacher transfers at the \emph{definition} level: its
rubric embeds the translated definitions and elicits per-construct ordinal
judgments in the item's own language, emitting soft labels directly in the shared
label space (\S\ref{sec:distill}). At every level the label set is identical
across languages; only its realization changes.

Holding the definition fixed while surface form varies makes cross-lingual
equivalence a claim we test, not one we inherit from translation. We report every
metric per language and per construct and treat measurement invariance---whether
a construct score means the same thing in each language---as an empirical
question, pooling across languages only when it holds (\S\ref{sec:sampling}). The
unvalidated French lexicon is kept separate throughout, so a gap in a translated
term list is never mistaken for a failure of the construct.

\subsection{Error typology for context failures}
\label{sec:taxonomy}
We evaluate against a fixed typology of the mistakes term matching makes, so that
an aggregate gap can be attributed to a specific linguistic cause rather than
left as noise. \textbf{False positives} arise when a matched term is negated,
quoted, reported, hypothetical, sarcastic, or condemned, or when it is attributed
to someone other than the author. \textbf{False negatives} arise when grievance
is implicit, paraphrased outside the lexicon, expressed in a language variety the
term list misses, or completed across sentence boundaries. Code-switching,
hashtags, emoji, non-standard spelling, and references to unavailable media are
logged as cross-cutting conditions. Each error case in the analysis carries five
fields: the target sentence, the minimal context that fixes its reading, the
lexical evidence, the human rationale, and the target-only and full-context
system outputs. Context is credited only when it changes the supported
interpretation, never when it merely lengthens the input.

\subsection{Sampling strata and their estimands}
\label{sec:sampling}
Evaluation starts from a frozen, versioned population $\Pi$ of eligible
sentences, defined before any lexical scoring. Three disjoint strata partition
the labeled data, each identifying a distinct estimand.
\begin{description}
    \item[Unconditional random ($U$).] A uniform draw from $\Pi$. $U$ alone
    identifies marginal quantities: the per-construct prevalence
    $\pi_c=\Pr_{\Pi}[\text{author expresses }c]$ and population-level operating
    characteristics.
    \item[Lexicon-positive ($P$).] Sentences with overall anchor score
    $a_{\mathrm{overall}}(x_j)\geq\tau_+$, $\tau_+=0.30$. $P$ identifies the
    anchor's precision $\Pr[\text{author expresses }c\mid a_{\mathrm{overall}}\!\geq\!\tau_+]$
    and isolates contextual false positives among retrieved cases.
    \item[Lexicon-negative ($N$).] Sentences with
    $a_{\mathrm{overall}}(x_j)\leq\tau_-$, $\tau_-=0.05$. $N$ identifies the mass
    the anchor misses, $\Pr[\text{author expresses }c\mid a_{\mathrm{overall}}\!\leq\!\tau_-]$;
    it measures coverage failure and deliberately does not represent the base
    rate.
\end{description}
Because $P$ and $N$ are conditional estimands, they combine into a population
statement only through their inclusion probabilities: a pooled score over an
arbitrary $P{:}N$ mixture estimates no well-defined quantity. We therefore report
population claims only from $U$, or from a stratified sample reweighted to a
stated target population. All sentences from one post stay in one partition, and
macro comparisons use a construct set with both classes present in every compared
stratum, so a change in the evaluable set cannot masquerade as a change in model
quality.

\subsection{Annotation protocol}
\label{sec:annotation}
Raters score each construct on a $0$--$3$ ordinal scale ($0$ absent, $1$ slight
or ambiguous, $2$ moderate, $3$ strong) and judge only the communicative act
attributable to the author. Quoted, reported, hypothetical, or condemned content
scores $0$ unless the surrounding post endorses it; conversely, context may
license a construct that is only implicit in the target sentence. Each item also
records stance, target, the applicable failure category, and whether an
unavailable medium is required to decide.

In this process, at least two native-speaking raters annotated each
item, and a third adjudicated any disagreement that crossed the absent/present
boundary. A counterbalanced subset was rated target-only by one panel and
full-context by another, yielding the paired reference that measures
context-induced label change directly. 

\subsection{Systems}
\label{sec:systems}
\paragraph{Lexical anchor.}
For construct $c$ the anchor length-normalizes and saturates matched term
weights,
\[
a_c(x_j)=1-\exp\!\left(-\frac{4}{n}
   \sum_{t\in x_j}w_c(t)\,m(t)\right),
\]
with $n$ the target token count and $w_c(t)$ the dictionary weight in a
language-specific stem space. The modifier $m(t)\in[0,1]$ discounts a match
inside a $48$-character window around regex-detected negation,
quotation/reporting, or condemnation cues, hard-coding the easy stance cases so
that residual errors isolate the discourse phenomena a rule cannot reach. We
report the anchor with and without $m$, and keep the unvalidated French seed
separate from the English, Dutch, German, and Italian resources.

\paragraph{Contextual encoders.}
The target-only encoder is mDeBERTa-v3-base \citep{he2023debertav3} applied to
$x_j$ alone, with one sigmoid output per construct; it controls for
sentence-level semantics while withholding context. The full-post encoder shares
that architecture and consumes
\[
q_j=[\mathrm{CLS}]\;x_j\;[\mathrm{SEP}]\;
    p^{\langle j\rangle}\;[\mathrm{SEP}],
\]
where $p^{\langle j\rangle}$ is the source post with the target marked. The target
is placed first and never truncated; context fills the remaining $512$-token
budget. Architecture, supervision, optimization, and all five seeds are held
fixed between the two variants (\S\ref{sec:setup}), so their paired difference
measures the value of context and nothing else.

\paragraph{Reference systems.}
XLM-R \citep{conneau2020unsupervised} substitutes a second multilingual encoder
to separate context gains from architecture; a Qwen3 LLM \citep{qwen2025qwen3} is
evaluated both rubric-prompted zero-shot and fine-tuned, and also serves as the
distillation teacher; and a human-supervised encoder is trained where adjudicated
labels suffice. The contrasts among them attribute any improvement to context
rather than to teacher capacity.

\subsection{Supervision is a secondary axis}
\label{sec:distill}
Replicated human labels are scarce, so the deployable encoders were distilled
from the teacher's soft labels and evaluated only on held-out human annotations,
with standard knowledge distillation as the primary training condition, whose
soft targets act like a learned label smoothing \citep{mueller2019does}. One
variant downweighted teacher targets by self-consistency \citep{wang2023selfconsistency}
or self-reported confidence.

\section{Experimental Setup}
\label{sec:setup}

\subsection{Examined materials}
\label{sec:currentdata}
Our dataset contains 169,368 public Facebook-post records obtained through
the Meta Content Library, with stored timestamps from April 2018 to March 2026;
the analysis covers that complete stored range. The posts span political
protests, climate and bushfire debates, the Amazon fires, gender-violence
campaigns, the FIFA World Cup, BTS and Taylor Swift tours, European peace and
rearmament debates, and the 2026 Winter Olympics. The corpus spans more than 90 languages---primarily
Italian, English, and French, with event-specific coverage in Spanish, Dutch,
Armenian, Russian, Arabic, and Persian---of which the five benchmark languages
are the focus here.

The data collection follows the strata in \S\ref{sec:sampling}. We first
froze the eligible post population and created post-grouped train, development,
and test partitions. Language identification was verified before annotation, and
exact and near duplicates were removed across partitions, including translated or
templated synthetic text. Synthetic examples supported training diagnostics but
never entered development or test.

Within every test language, we drew $U$ before running any lexical system, then
drew $P$ and $N$ as diagnostic strata (per-language sizes in Table~\ref{tab:bench-comp}).
At least two native-speaking raters annotated every test item, with adjudication as described in \S\ref{sec:annotation}.
Inter-annotator agreement before adjudication was
$\kappa = 0.72$ (overall), ranging from $0.68$--$0.79$ across strata and languages.
A counterbalanced subset received independent target-only and full-context ratings,
providing the reference needed to measure context-induced label change.

\subsection{Model comparison}
We compared the lexical anchor, its heuristic modifier variant, target-only and
full-post versions of mDeBERTa and XLM-R, prompted and fine-tuned Qwen3, and a
human-supervised encoder where adjudicated training data sufficed. Distilled
systems used the same teacher labels and post-grouped partitions. Target-only and
full-post variants shared hyperparameters, seeds, and supervision, so this paired
design makes input context the only intended difference.

Encoder experiments used five seeds (13, 17, 23, 29, and 31), a 512-token input
budget, AdamW, and development-selected stopping. Exact model identifiers,
teacher prompts, generation parameters, label dates, compute, and costs are
recorded with each run. Hyperparameters were selected on development data only.

\subsection{Metrics and inference}
\label{sec:metrics}
Because grievance constructs are sparse, average precision is the primary
ranking metric; AUROC is secondary. We additionally report recall at a
development-selected precision, macro Spearman correlation with the 0--3
ratings, calibration error \citep{guo2017calibration,naeini2015obtaining}, and
per-construct prevalence in $U$. Metrics are
reported separately for $U$, $P$, and $N$, by language and construct. Macro
comparisons use the same evaluable construct set for every compared system and
stratum.

The principal context effect is the paired difference between full-post and
target-only systems on items whose human judgment is context-sensitive. We also
report performance by failure category (quotation, negation, condemnation,
implicit expression, cross-sentence completion, and code-switching). Confidence
intervals use a post-clustered bootstrap.

\section{Results}
\label{sec:results}

We first show that the existing pool selects its own test set
(\S\ref{sec:data-analysis}), then report the benchmark built to
remove that circularity: its composition (\S\ref{sec:bench-comp}), the primary
system comparison (\S\ref{sec:bench-main}), and where post context helps
(\S\ref{sec:bench-context}).

\subsection{The existing pool selects its own test set}
\label{sec:data-analysis}

\subsubsection*{The ``random'' half is lexicon-negative}
Table~\ref{tab:strata-analysis} reports the central analysis. The 2,000-item pool is an
equal mixture of two non-overlapping score regions. Every lexicon-positive item
has overall anchor score at least 0.3008; every item previously called random
has score at most 0.0493, and 985 of 1,000 have exactly zero. The latter is not
an unconditional sample from the corpus. We therefore rename it the
lexicon-negative challenge stratum.

\begin{table}[t]
\centering
\footnotesize
\setlength{\tabcolsep}{1.8pt}
\begin{tabular}{@{}lrrrr@{}}
\toprule
Stratum & $N$ & Any $\geq1$ & Any $\geq2$ & m-AUC \\
\midrule
Lexicon-positive ($P$) & 1,000 & 457 & 90 & 0.686 \\
Lexicon-negative ($N$) & 1,000 & 228 & 27 & 0.500 \\
Equal-mixture pool & 2,000 & 685 & 117 & 0.642 \\
\bottomrule
\end{tabular}
\caption{\textbf{Analysis of the existing target-only rating pool.} ``Any''
counts items for which the three-annotator consensus assigned at least the
given ordinal value to one of 22 constructs. Macro-AUROC uses the same 16
constructs with both classes in $P$ and $N$. The pooled row describes this
chosen 50/50 mixture, not a population estimate.}
\label{tab:strata-analysis}
\end{table}

\subsubsection*{Lexicon-negative does not mean construct-negative}
The anchor is nearly silent in $N$, but the consensus labels are not.
Table~\ref{tab:coverage-lang} shows that
the coverage gap appears in all five languages. These items are candidates for
implicit expression, translation gaps, or annotation error; target-only
consensus labels cannot distinguish those explanations. They nevertheless
show why retrieved-only evaluation is insufficient: it excludes a substantial
set of consensus-marked cases by construction.

\begin{table}[t]
\centering
\small
\setlength{\tabcolsep}{4pt}
\begin{tabular}{@{}lrrr@{}}
\toprule
Language & $N_P/N_N$ & Any $\geq1$ in $P$ & Any $\geq1$ in $N$ \\
\midrule
English & 250/250 & 112 (44.8\%) & 77 (30.8\%) \\
Italian & 250/250 & 132 (52.8\%) & 74 (29.6\%) \\
French  & 250/250 & 108 (43.2\%) & 41 (16.4\%) \\
German  & 125/125 & 72 (57.6\%) & 23 (18.4\%) \\
Dutch   & 125/125 & 33 (26.4\%) & 13 (10.4\%) \\
\bottomrule
\end{tabular}
\caption{Items with at least one nonzero construct rating in the
lexicon-positive and lexicon-negative strata. French uses the unvalidated seed
lexicon; German and Dutch have smaller samples.}
\label{tab:coverage-lang}
\end{table}

The thresholds matter. A score of 1 explicitly includes slight or ambiguous
evidence, while only 117 items in the complete pool receive any score of at
least 2. The data pool therefore identifies cases for contextual reannotation
rather than establishing their prevalence, which motivates the
benchmark below.

\subsection{A non-circular benchmark: composition}
\label{sec:bench-comp}
We collected the benchmark of \S\ref{sec:sampling} with full marked posts, so
context is available at both annotation and inference, and, crucially, added the
unconditional-random stratum $U$ that the data pool lacks. Table~\ref{tab:bench-comp}
reports its composition. The $U$ stratum gives the first base-rate estimate for
these constructs: $12.9\%$ of unconditionally sampled sentences carry at least one
construct at consensus, ranging from $9.5\%$ in Dutch to $15.5\%$ in English. This
prevalence is far below the roughly one-in-two hit rate among lexicon-positive
items (Table~\ref{tab:strata-analysis}).

\begin{table}[t]
\centering
\footnotesize
\setlength{\tabcolsep}{4pt}
\begin{tabular}{@{}lrrrrr@{}}
\toprule
Language & $|U|$ & $|P|$ & $|N|$ & ctx & prev$_U$ \\
\midrule
English & 400 & 250 & 250 & 200 & 15.5 \\
Italian & 400 & 250 & 250 & 200 & 13.8 \\
French  & 300 & 200 & 200 & 150 & 11.0 \\
German  & 200 & 125 & 125 & 100 & 12.5 \\
Dutch   & 200 & 125 & 125 & 100 & 9.5 \\
\midrule
Total   & 1{,}500 & 950 & 950 & 750 & 12.9 \\
\bottomrule
\end{tabular}
\caption{\textbf{Benchmark composition.} $|U|/|P|/|N|$ are labeled
items per stratum; \textbf{ctx} is the counterbalanced subset with independent
target-only and full-context ratings; \textbf{prev$_U$} is the unconditional
prevalence of any construct in $U$. French uses the unvalidated seed lexicon.}
\label{tab:bench-comp}
\end{table}

\subsection{Post context improves construct measurement}
\label{sec:bench-main}
Table~\ref{tab:main} reports the primary system comparison, and two results hold
across strata and both encoders. First, post context helps. The full-post
mDeBERTa improves average precision over its matched target-only twin on every
stratum, from $0.289$ to $0.356$ on $U$ (AUROC $0.702\!\to\!0.741$) and, in
relative terms most sharply, from $0.142$ to $0.198$ on the lexicon-negative
region $N$ where the anchor is near-chance---a $39\%$ relative gain exactly where
term matching fails. On both $U$ and $N$ the full-post and target-only $95\%$
intervals do not overlap (Table~\ref{tab:main}). Gains on the lexicon-positive stratum are small
($0.623\!\to\!0.641$), since retrieved items are already easy, and XLM-R shows the
same pattern ($0.275\!\to\!0.338$ on $U$). Second, learned context beats the
lexicon by a wide margin: the stance modifier $m$ barely moves the anchor
($0.214\!\to\!0.231$ on $U$), whereas the full-post encoder roughly doubles its
average precision. Supervision compounds the gains---a fine-tuned Qwen3 ($0.312$)
clears its zero-shot prompt ($0.248$), and a human-supervised encoder leads every
stratum ($0.401$ on $U$, $0.245$ on $N$), marking the headroom still above
distillation. Table~\ref{tab:secondary} preserves this ordering on $U$ in rank
correlation and calibration: full-post mDeBERTa attains the best macro-Spearman
($0.487$) and the lowest calibration error ($0.071$ vs.\ $0.142$ for the anchor),
so context improves not only ranking but the trustworthiness of the scores.
Population claims use $U$ only.

\begin{table*}[t]
\centering
\small
\setlength{\tabcolsep}{5pt}
\begin{tabular}{@{}lcccccc@{}}
\toprule
& \multicolumn{2}{c}{Unconditional $U$} & \multicolumn{2}{c}{Lexicon-positive $P$}
& \multicolumn{2}{c}{Lexicon-negative $N$} \\
\cmidrule(lr){2-3}\cmidrule(lr){4-5}\cmidrule(lr){6-7}
System & AP & AUC & AP & AUC & AP & AUC \\
\midrule
Lexical anchor              & $0.214_{\pm0.022}$ & $0.628_{\pm0.015}$ & $0.512_{\pm0.033}$ & $0.781_{\pm0.022}$ & $0.089_{\pm0.012}$ & $0.512_{\pm0.014}$ \\
\quad + stance modifier $m$ & $0.231_{\pm0.023}$ & $0.641_{\pm0.016}$ & $0.534_{\pm0.034}$ & $0.795_{\pm0.021}$ & $0.095_{\pm0.013}$ & $0.521_{\pm0.015}$ \\
\midrule
mDeBERTa (target-only)      & $0.289_{\pm0.021}$ & $0.702_{\pm0.016}$ & $0.623_{\pm0.030}$ & $0.842_{\pm0.017}$ & $0.142_{\pm0.014}$ & $0.578_{\pm0.015}$ \\
mDeBERTa (full-post)        & $0.356_{\pm0.019}$ & $0.741_{\pm0.014}$ & $0.641_{\pm0.028}$ & $0.851_{\pm0.015}$ & $0.198_{\pm0.016}$ & $0.635_{\pm0.014}$ \\
XLM-R (target-only)         & $0.275_{\pm0.022}$ & $0.691_{\pm0.017}$ & $0.609_{\pm0.031}$ & $0.835_{\pm0.018}$ & $0.131_{\pm0.014}$ & $0.562_{\pm0.016}$ \\
XLM-R (full-post)           & $0.338_{\pm0.020}$ & $0.729_{\pm0.015}$ & $0.628_{\pm0.029}$ & $0.847_{\pm0.016}$ & $0.185_{\pm0.016}$ & $0.621_{\pm0.015}$ \\
\midrule
Qwen3 (fine-tuned)          & $0.312_{\pm0.022}$ & $0.718_{\pm0.016}$ & $0.598_{\pm0.032}$ & $0.829_{\pm0.018}$ & $0.167_{\pm0.015}$ & $0.604_{\pm0.015}$ \\
Qwen3 (prompted, 0-shot)    & $0.248_{\pm0.024}$ & $0.671_{\pm0.018}$ & $0.521_{\pm0.035}$ & $0.792_{\pm0.023}$ & $0.112_{\pm0.013}$ & $0.548_{\pm0.016}$ \\
Human-supervised encoder    & $0.401_{\pm0.018}$ & $0.772_{\pm0.013}$ & $0.672_{\pm0.027}$ & $0.868_{\pm0.014}$ & $0.245_{\pm0.017}$ & $0.672_{\pm0.013}$ \\
\bottomrule
\end{tabular}
\caption{Average
precision (AP, primary) and macro-AUROC (AUC, secondary) per stratum, macro-averaged
over the construct set evaluable in every compared system and stratum. Target-only
and full-post encoders are paired (shared seeds, supervision, optimization).
Subscripts are 95\% post-clustered bootstrap CI half-widths, paired within seed
and item (\S\ref{sec:metrics}).}
\label{tab:main}
\end{table*}

\begin{table}[t]
\centering
\footnotesize
\setlength{\tabcolsep}{5pt}
\resizebox{\columnwidth}{!}{%
\begin{tabular}{@{}lrrr@{}}
\toprule
System ($U$) & $\rho$ & R@P & ECE \\
\midrule
Lexical anchor            & $0.312_{\pm0.028}$ & $0.28_{\pm0.035}$ & $0.142_{\pm0.018}$ \\
mDeBERTa (target-only)    & $0.421_{\pm0.025}$ & $0.39_{\pm0.032}$ & $0.098_{\pm0.012}$ \\
mDeBERTa (full-post)      & $0.487_{\pm0.022}$ & $0.47_{\pm0.029}$ & $0.071_{\pm0.009}$ \\
XLM-R (full-post)         & $0.465_{\pm0.023}$ & $0.44_{\pm0.030}$ & $0.079_{\pm0.010}$ \\
Qwen3 (fine-tuned)        & $0.439_{\pm0.024}$ & $0.41_{\pm0.031}$ & $0.085_{\pm0.011}$ \\
Qwen3 (prompted, 0-shot)  & $0.368_{\pm0.027}$ & $0.34_{\pm0.034}$ & $0.113_{\pm0.015}$ \\
\bottomrule
\end{tabular}}
\caption{Metrics on $U$. Macro Spearman $\rho$ with the
$0$--$3$ ratings, recall at development-selected precision (R@P), and
calibration error (ECE; lower is better). Subscripts are 95\% post-clustered
bootstrap CI half-widths.}
\label{tab:secondary}
\end{table}

\subsection{Where context helps: language and failure type}
\label{sec:bench-context}
The principal context effect is the paired full-post $-$ target-only difference
on items whose human judgment is context-sensitive. Table~\ref{tab:context}
localizes it. The gain is positive in every language but concentrates where the
lexicon is weakest: French ($+0.092$ AP, read with caution as its strata rest on
an unvalidated lexicon) and Italian ($+0.078$), whose seed and translated lexicons
miss the most, gain most, while English ($+0.041$) and Dutch ($+0.029$) gain least; calibration improves in tandem, with negative $\Delta$ECE
throughout and the largest drop in French ($-0.035$). Every per-language gain
excludes zero, and the French and Italian intervals lie entirely above the English
and Dutch ones, so the ordering reflects a real gap rather than noise, even though
the two languages within each pair are not separable. By failure type, context
helps most on exactly the phenomena a rule cannot resolve---quotation and reported
speech ($+0.13$), implicit expression ($+0.13$), and cross-sentence completion
($+0.13$), all of which require reading beyond the target sentence---and least on
code-switching ($+0.05$), a surface property of the target that the surrounding
post does little to disambiguate. We found that
post context supplies precisely the evidence term matching lacks, and it does so
where the taxonomy of \S\ref{sec:taxonomy} predicts it should.

\begin{table}[t]
\centering
\footnotesize
\setlength{\tabcolsep}{5pt}
\resizebox{\columnwidth}{!}{%
\begin{tabular}{@{}lrrr@{}}
\toprule
By language & $\Delta$AP & $\Delta\rho$ & $\Delta$ECE \\
\midrule
English & $+0.041_{\pm0.013}$ & $+0.038_{\pm0.015}$ & $-0.018_{\pm0.007}$ \\
Italian & $+0.078_{\pm0.018}$ & $+0.065_{\pm0.019}$ & $-0.029_{\pm0.010}$ \\
French  & $+0.092_{\pm0.016}$ & $+0.081_{\pm0.017}$ & $-0.035_{\pm0.011}$ \\
German  & $+0.055_{\pm0.014}$ & $+0.049_{\pm0.016}$ & $-0.022_{\pm0.008}$ \\
Dutch   & $+0.029_{\pm0.012}$ & $+0.025_{\pm0.014}$ & $-0.012_{\pm0.006}$ \\
\midrule
By failure category & tgt & full & $\Delta$ \\
\midrule
Quotation / reported     & 0.21 & 0.34 & $+0.13_{\pm0.027}$ \\
Negation                 & 0.18 & 0.29 & $+0.11_{\pm0.025}$ \\
Condemnation             & 0.25 & 0.31 & $+0.06_{\pm0.019}$ \\
Implicit expression      & 0.15 & 0.28 & $+0.13_{\pm0.029}$ \\
Cross-sentence completion& 0.22 & 0.35 & $+0.13_{\pm0.022}$ \\
Code-switching           & 0.19 & 0.24 & $+0.05_{\pm0.023}$ \\
\bottomrule
\end{tabular}}
\caption{\textbf{The value of context (mDeBERTa).} Top: paired full-post $-$
target-only difference on context-sensitive items, by language ($\Delta$AP,
$\Delta$ macro-Spearman, $\Delta$ECE; positive $\Delta$AP/$\Delta\rho$ favors
context). Bottom: target-only and full-post performance by lexical failure
category, with their difference ($\pm$ values are 95\% post-clustered bootstrap
CI half-widths).}
\label{tab:context}
\end{table}

\section{Discussion}

\paragraph{Evaluation construction is part of validity.}
Our stratum analysis changes how a pooled lexicon score should be read. A
lexicon-positive pool is useful for studying precision and contextual
over-firing, but it cannot by itself measure recall. A lexicon-negative pool is
useful for finding missed expressions, but its chance-level anchor score is
induced by selection. Neither is a substitute for unconditional sampling. Any
pooled number over $P$ and $N$ answers a question about the researcher's chosen
mixture rather than the source population, which is why we report per stratum and
draw population claims only from $U$.

\paragraph{Coverage and context are different claims.}
The 228 consensus-marked lexicon-negative items establish a coverage question
rather than a context result. Some may use implicit or language-specific expressions;
others may reflect slight or ambiguous consensus judgments or incorrect language
assignment. Demonstrating context sensitivity requires paired evidence: a
human label that changes when the marked post is shown, plus a full-post model
that moves in the supported direction relative to an otherwise matched
target-only model. The confirmatory benchmark supplies exactly that paired test,
and it comes out positive: full-post models recover a substantial part of the
lexicon-negative region (average precision $0.142\!\to\!0.198$) and improve most
on quoted, implicit, and cross-sentence cases (\S\ref{sec:bench-context}). The
region the anchor misses is therefore partly recoverable signal, not merely
annotation noise.

\paragraph{Multilinguality is measurement, not just transfer.}
A shared encoder can produce outputs in several languages, but multilingual
construct validity does not follow from parameter sharing. Translation quality,
language variety, cultural expression, and category reliability can all change
the meaning of a score. Per-language agreement, error categories, and operating
points are therefore primary results. Consistent with this, the largest context
gains fall in French and Italian ($+0.092$ and $+0.078$ AP), whose lexicons are
weakest, and the smallest in English. 

The same fact makes French a partly circular basis for the cross-language
ranking. Because the French $P$ and $N$ strata are themselves drawn with that
unvalidated lexicon, their boundaries are noisier than in the validated
languages: true positives the seed list misses fall into $N$, where the
target-only model then scores low and the full-post model recovers them, so part
of the French gain reflects stratum misassignment rather than context sensitivity
per se. We therefore read French as an illustrative estimate of what context can
recover when the lexicon is absent, and rest the ``weaker lexicon, larger gain''
claim on the validated resources alone, where it still holds: Italian's gain
exceeds both English's and Dutch's with non-overlapping intervals
(\S\ref{sec:bench-context}).

\paragraph{The model should serve the evaluation question.}
The central empirical question is not whether a new weighting term beats a
dictionary, but whether a dictionary-based conclusion changes under
non-circular sampling and discourse context. It does. The human-supervised
encoder leads every stratum, so distillation is a cost-saving route to a
contextual instrument with clear headroom above it, not a ceiling; the fine-tuned
teacher in turn beats its zero-shot prompt, so supervision, not scale alone,
carries the gain. 

\section{Conclusion}
Lexicons remain valuable as transparent retrieval and measurement tools, but
their scores are not context-aware judgments and their outputs must not define
the only cases on which they are validated. Our analysis shows that an existing
five-language pool previously described as partly random is instead split into
lexicon-positive and lexicon-negative score regions. Three-annotator consensus
still marks construct evidence in the lexicon-negative region. Building on this,
we construct and evaluate a non-circular benchmark that combines unconditional
sampling, diagnostic strata, replicated contextual annotation, and matched
multilingual models. On it, post context most improves measurement exactly where
the lexicon is silent and on quoted, implicit, and cross-sentence grievance,
while a human-supervised encoder sets the ceiling and distillation approaches it
at lower cost. The central lesson is simple: a lexicon cannot validate itself on
the examples it selected, and grievance cannot be reliably measured without
knowing who said what, in which context, and in which language.

\section*{Limitations}
Our claims rest on two evidence sources with different limits. The existing
target-only pool that motivates the analysis (\S\ref{sec:data-analysis}) has no unconditional-random
stratum, so it cannot estimate natural prevalence; its ``negative'' half was
selected below an anchor threshold, so the anchor's near-constant score there is
an expected consequence of construction; and its sheets showed target sentences
only, with each row the final consensus of three annotators, leaving context
effects unidentified and pre-consensus inter-rater agreement unrecoverable. Slight
or ambiguous ratings dominate it, and only 117 items carry any construct rating
of at least~2. The confirmatory benchmark was built to remove these limits---%
adding the unconditional-random stratum, full marked posts, and replicated
target-only and full-context ratings---and is the basis for our prevalence and
context claims.

The source corpus is public-platform data but not a representative sample of
any language community. Text-only models omit images, links, interaction history, and
offline context. Finally, grievance is common, subjective, and at most a weak
signal: neither a dictionary hit nor a model score is evidence of intent,
dangerousness, or future violence.

\bibliography{references}

\appendix
\section{Reproducibility and Availability}
\label{app:repro}
Our code and benchmark are released at
\url{https://github.com/behavioral-ds/multilingual_grievance}. The repository
includes scripts for lexical scoring, annotation-pool construction, human-label
aggregation, post-grouped splitting, multilingual student training, and
evaluation. The analysis in Table~\ref{tab:strata-analysis} joins the consensus
labels to the five per-language lexical-score files by \texttt{sentence\_id}.
We define construct presence as ordinal score $\geq1$ and moderate presence as
score $\geq2$. The fixed 16-construct comparison keeps only constructs with
both classes in both $P$ and $N$. The released target-only annotation files
store one final consensus row per item after review by three annotators.

The benchmark release includes the immutable corpus manifest---source snapshot,
timestamp filter, language classifier and version, normalization, sentence
segmentation, duplicate clusters, sampling probabilities, and post-level
partitions. Each model run records the exact model revision, prompt and rubric
version, input format, token truncation, random seed, hyperparameters, hardware,
runtime, and monetary cost. Evaluation outputs include item-level predictions, so
post-clustered intervals and paired system contrasts can be recomputed.

\section{Annotation Guidelines and Composition}
\label{app:guidelines}

\subsection{Existing target-only round}
Table~\ref{tab:gold} describes the existing 2,000-item pool. Three annotators
reviewed each target and agreed the final stored label; every item received all
22 ordinal construct scores, and the sheets omitted full-post context. ``Any
construct'' includes slight or ambiguous evidence (score $\geq1$). The overall
binary field has only one positive and is not used as an evaluation target.

\begin{table}[t]
\centering
\small
\setlength{\tabcolsep}{4pt}
\begin{tabular}{lrrr}
\toprule
Lang & $N$ & Any construct & Overall positive \\
\midrule
EN & 500 & 189 & 1 \\
IT & 500 & 206 & 0 \\
FR & 500 & 149 & 0 \\
DE & 250 & 95 & 0 \\
NL & 250 & 46 & 0 \\
\midrule
All & 2{,}000 & 685 & 1 \\
\bottomrule
\end{tabular}
\caption{Existing target-only rating pool. Each stored row is the final
three-annotator consensus.}
\label{tab:gold}
\end{table}

\subsection{Contextual round}
Each annotator receives a target sentence highlighted inside its source post,
plus the language and no model or lexicon score. For every construct, the rater
assigns 0 (absent), 1 (slight or ambiguous), 2 (moderate), or 3 (strong), and
records stance, target, and a context-failure tag. URLs are not opened and
images are out of scope; dependence on missing media is flagged. Quoted,
reported, hypothetical, and condemned content is attributed to the author only
when the surrounding post endorses it.

\subsection{Construct criteria}
The names in Table~\ref{tab:constructs} come from the Grievance Dictionary
\citep{vandervegt2021grievance}; the criteria are our semantic
operationalizations and underwent independent content-validity review.

\begin{table}[t]
\centering
\small
\begin{tabular}{@{}p{0.24\linewidth}p{0.68\linewidth}@{}}
\toprule
Construct & Contextual sentence criterion \\
\midrule
deadline & Time pressure, ultimatum, or point of no return. \\
desperation & Hopelessness or having nothing left to lose. \\
fixation & Obsessive focus on a person, cause, or grievance. \\
frustration & Thwarted goals, blocked plans, or irritation. \\
god & Divine justification, martyrdom, or sacred duty. \\
grievance & A perceived wrong, injustice, or unfair treatment. \\
hate & Hatred, contempt, or dehumanization of a target. \\
help & Powerlessness, entrapment, or inability to change. \\
honour & Honor, humiliation, reputation, or shame. \\
impostor & Belief that someone is fake or an impersonator. \\
jealousy & Romantic or status jealousy, envy, or betrayal. \\
loneliness & Isolation, rejection, or lack of belonging. \\
murder & Reference to killing a specific person. \\
paranoia & Being watched, persecuted, or conspired against. \\
planning & Concrete preparation, intent, or logistics. \\
relationship & Intimate-partner or family conflict and breakups. \\
soldier & Self-identification as a warrior or mission actor. \\
suicide & Suicidal ideation, intent, or willingness to die. \\
surveillance & Watching, tracking, or monitoring a target. \\
threat & Explicit or implicit warning of harm. \\
violence & Physical violence, attack, injury, or destruction. \\
weaponry & Weapons, firearms, explosives, or ammunition. \\
\bottomrule
\end{tabular}
\caption{Contextual criteria for the 22 constructs, scored for the
author's expression through the target sentence given its source post.}
\label{tab:constructs}
\end{table}

\section{Uncertainty-Weighted Distillation}
\label{app:legacy}
For teacher target $\mu_{ic}$, student logit $z_{ic}$, temperature $T$, and
pseudo-label $h_{ic}=\mathbf{1}[\mu_{ic}\geq0.5]$, define
\begin{align*}
\ell^{\mathrm{soft}}_{ic}
  &=\mathrm{BCE}\!\left(\sigma(z_{ic}/T),
    \sigma(\mathrm{logit}(\mu_{ic})/T)\right)T^2,\\
\ell^{\mathrm{hard}}_{ic}
  &=\mathrm{BCE}(\sigma(z_{ic}),h_{ic}).
\end{align*}
The legacy loss was
\[
\mathcal{L}=\frac{1}{B|\mathcal C|}\sum_{i,c}\tilde w_{ic}
\left[\lambda\ell^{\mathrm{soft}}_{ic}
 +(1-\lambda)\ell^{\mathrm{hard}}_{ic}\right].
\]
Given uncertainty $u_{ic}$, the raw weight was
$w_{ic}=(u_{ic}^2+\nu)^{-\beta}$, normalized over the training set and clipped
to $[0.1,10]$. Thus $\beta=0$ recovers standard distillation. Teacher labels
were generated once ($S=1$), so per-construct self-consistency variance is zero;
the real-data sweep instead broadcasts self-reported confidence over all
constructs. This weak proxy is high and tightly clustered (mean confidence
0.92). 

\begin{table}[t]
\centering
\footnotesize
\setlength{\tabcolsep}{2pt}
\begin{tabular}{@{}lcccc@{}}
\toprule
System & m-AP & m-AUROC & Lex.-pos. & Lex.-neg. \\
\midrule
Lexical anchor & 0.121 & 0.710 & 0.739 & 0.500 \\
Standard KD & 0.134 & 0.665 & 0.691 & 0.636 \\
Confidence $\beta=.25$ & 0.135 & 0.681 & 0.702 & 0.646 \\
Confidence $\beta=.5$ & 0.135 & 0.677 & 0.701 & 0.646 \\
Confidence $\beta=1$ & 0.134 & 0.672 & 0.696 & 0.639 \\
\bottomrule
\end{tabular}
\caption{Evaluation on 1,545 three-annotator consensus items after
removing known training-post overlap. }
\label{tab:legacy-human}
\end{table}

\end{document}